\documentclass{article}

\usepackage{microtype}
\usepackage{graphicx}
\usepackage{multicol}
\usepackage{subfigure}
\usepackage{booktabs} 
\usepackage{adjustbox}

\usepackage{hyperref}

\usepackage[accepted]{icml2024}

\usepackage{amsmath}
\usepackage{amssymb}
\usepackage{mathtools}
\usepackage{amsthm}
\usepackage{booktabs} 
\usepackage{tabularx} 

\usepackage[capitalize,noabbrev]{cleveref}

\theoremstyle{plain}

\theoremstyle{definition}

\theoremstyle{remark}

\usepackage[textsize=tiny]{todonotes}

\icmltitlerunning{Enhancing Antibiotic Stewardship using a Natural Language Approach for Better Feature Representation}

\begin{document}

\twocolumn[
\icmltitle{Enhancing Antibiotic Stewardship using a Natural Language Approach for Better Feature Representation}



\icmlsetsymbol{equal}{*}

\begin{icmlauthorlist}
\icmlauthor{Simon A. Lee}{yyy}
\icmlauthor{Trevor Brokowski}{zzz,aaa}
\icmlauthor{Jeffrey N. Chiang}{yyy,comp}
\end{icmlauthorlist}

\icmlaffiliation{yyy}{Department of Computational Medicine, UCLA}
\icmlaffiliation{comp}{Department of Neurosurgery, UCLA}
\icmlaffiliation{zzz}{Yale School of Medicine}
\icmlaffiliation{aaa}{Biomedical Informatics \& Data Science, Yale University}

\icmlcorrespondingauthor{Simon A. Lee}{simonlee711@g.ucla.edu}
\icmlcorrespondingauthor{Jeffrey Chiang}{njchiang@g.ucla.edu}

\icmlkeywords{Machine Learning, ICML}

\vskip 0.3in
]



\printAffiliationsAndNotice{\icmlEqualContribution} 

\begin{abstract}
The rapid emergence of antibiotic-resistant bacteria is recognized as a global healthcare crisis, undermining the efficacy of life-saving antibiotics. This crisis is driven by the improper and overuse of antibiotics, which escalates bacterial resistance. In response, this study explores the use of clinical decision support systems, enhanced through the integration of electronic health records (EHRs), to improve antibiotic stewardship. However, EHR systems present numerous data-level challenges, complicating the effective synthesis and utilization of data. In this work, we transform EHR data into a serialized textual representation and employ pretrained foundation models to demonstrate how this enhanced feature representation can aid in antibiotic susceptibility predictions. Our results suggest that this text representation, combined with foundation models, provides a valuable tool to increase interpretability and support antibiotic stewardship efforts.
\end{abstract}

\section{Introduction}

The Centers for Disease Control and Prevention (CDC) has declared the rapid emergence of resistant bacteria a global healthcare crisis, threatening the efficacy of antibiotics that have saved millions of lives \citep{ventola2015antibiotic, golkar2014bacteriophage, gould2013new, sengupta2013multifaceted, nature2013antibiotic, lushniak2014antibiotic}. This crisis is primarily driven by the mishandling and overuse of these antibiotics, which leads to bacteria developing resistance through repetitive exposure \cite{viswanathan2014off, read2014antibiotic}. 
These resistant bacteria impose significant clinical and financial burdens on healthcare systems, as well as on patients and their families worldwide \cite{bartlett2013seven}.

Clinical decision support systems hold substantial potential to assist healthcare providers in adhering to antibiotic stewardship practices. This potential is largely facilitated by electronic health record (EHR) software, which allows for the seamless integration of patient health histories in digital form \cite{evans2016electronic, cowie2017electronic, hoerbst2010electronic}. The integration with EHRs enables the use of continuously updated and deployed machine learning models for clinical decision-making. However, EHR data in its raw form presents numerous challenges, such as data ingestion and feature representation \cite{wu2010prediction}.

In this work, we present a methodology that converts EHR data into a serialized text form called pseudo-notes to predict antibiotic susceptibility. This conversion from tabular to text format facilitates the creation of interpretable data inputs for pretrained foundation models, known for their rich feature representation. Our primary objective is to develop a predictive model that incorporates this representation strategy in conjunction with foundation models. This approach aims to enhance decision support systems and accurately identify the most suitable antibiotics for patients, thereby offering a data-driven solution to combat antibiotic resistance.

\section{Related Works}

\paragraph{Medical Representation Learning}

Medical representation learning on Electronic Health Records (EHRs) has emerged as a critical area in healthcare research, focusing on transforming complex medical data for enhanced clinical decision-making. Initially, this involved extensive feature engineering to convert raw EHR data into formats suitable for traditional machine learning models \cite{tang2020democratizing, ferrao2016preprocessing}. However, this approach can be labor-intensive and varies significantly across research groups due to the lack of a standard protocol. 

Recently, the focus has shifted to advanced foundation models that learn to represent medical data by analyzing extensive text corpora, including clinical notes, medical literature, and records. These models, primarily based on the BERT architecture, generate rich, contextual latent representations of patient histories, significantly reducing the manual effort in feature engineering  \cite{rasmy2021med,liu2021med,alsentzer2019publicly,lee2020biobert}. Moreover, new techniques have been developed to incorporate the longitudinal nature of EHR data, leveraging a patients progression to predict outcomes \cite{steinberg2023motor, wornow2024ehrshot, pang2021cehr, li2022hi}.

\paragraph{Clinical Decision Support}

Machine learning enhances clinical decision support systems by analyzing vast datasets to provide evidence-based recommendations that improve patient care outcomes \cite{sutton2020overview}. Since the accessibility to EHR systems, there have been numerous such use cases in predicting diseases \cite{liu2018deep, cheng2016risk}, and various other patient outcomes \cite{lee2024emergency, suter1994predicting, churpek2014using} across all institutions within the healthcare system. Furthermore, the integration of predictive models into clinical workflows enables researchers to preemptively manage chronic conditions and mitigate potential health crises before they escalate \cite{li2020electronic, goldstein2017opportunities, hohman2023leveraging}. By continuously learning from new data, these systems evolve to provide more accurate assessments and recommendations, which support ongoing improvements in medical practices and patient management strategies. However, much work remains to be done on addressing the generalization and biases prevalent in many of these predictive algorithms \cite{goetz2024generalization, agniel2018biases}.

\section{Methods}
\paragraph{EHR} Electronic Health Records are digital versions of a patient’s medical history, maintained over time by healthcare providers. These records are valuable but consist of heterogeneous tabular datasets organized into separate tables such as diagnostics, demographics, and medication, presenting numerous challenges for researchers.

The primary challenge with EHRs is the heterogeneous nature of the data, which includes numerical, categorical, and free-text formats that are difficult to integrate and convert into machine-readable formats. Furthermore, categorical data fields often contain a large number of classes, potentially distorting their original representation. For instance, employing feature engineering techniques such as dummy coding for categorical variables can introduce collinearity, increase dimensionality, and result in sparse data representations. These modifications can complicate simple table readouts and require more memory capacity for statistical models to function effectively.

\paragraph{Pseudo-notes: Clinical Notes Generation from Tabular Data} Recent research has introduced a methodology for serializing tabular data into text using text templates \cite{hegselmann2023tabllm}. This approach significantly enhances our work by enabling uniform representation of all data in the EHR as human-readable and interpretable text, rather than as a collection of merged tables. Moreover, it creates an interface to use foundation models pre-trained on large text corpora, facilitating rich feature representation. In our work, we use a mapping function \( f: T \rightarrow S \) that turns individual tables into serialized text, where \( T \) stands for individual tables and \( S \) for serialized text. For each patient, we convert each of their $N$ tables—which cover different aspects like diagnostics, medications, and vitals—into text segments. These segments are then joined to create a single, detailed paragraph per patient. This method combines all pertinent patient information from various sources into one unified narrative, effectively transforming the data structure into \( S = \bigcup_{i=1}^{N} f(T_i) \), where \( T_i \) is the ith table row concerning the patient.

\paragraph{Data Source and Inclusion Criteria}

\begin{figure*}[t!]
\vskip -0.1in
\begin{center}
\centerline{\includegraphics[width=6.5in]{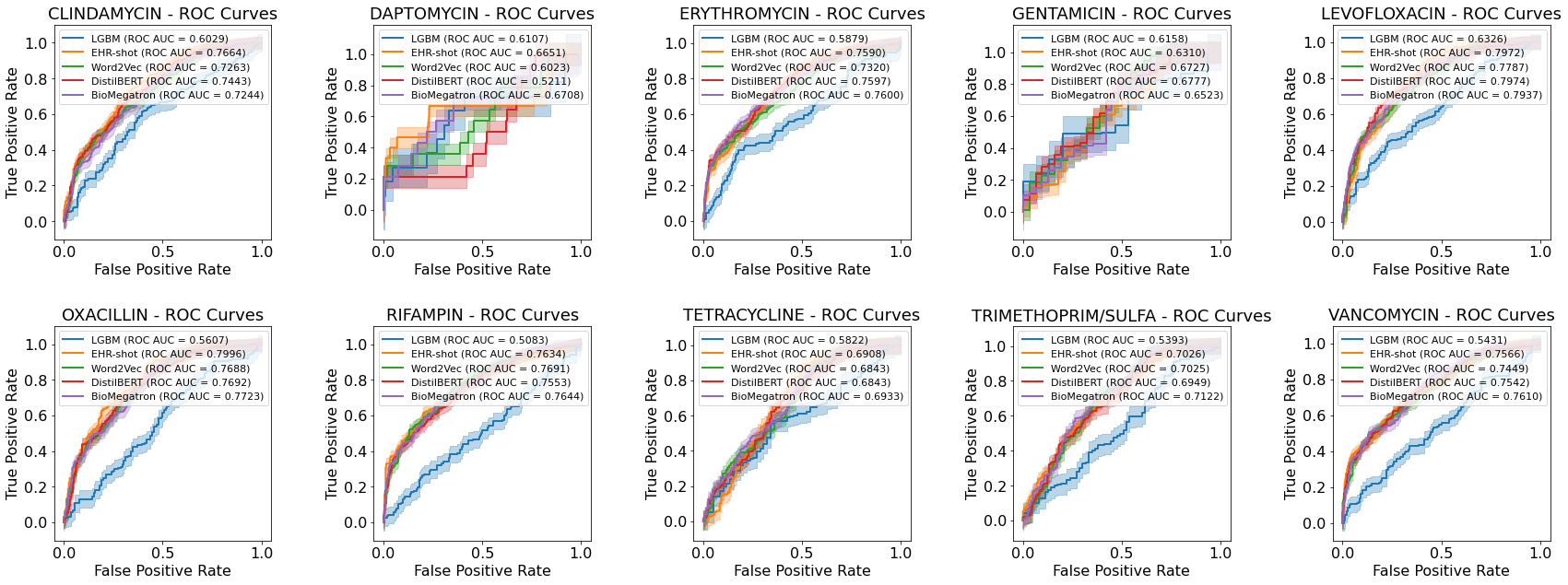}}
\caption{Area under the Receiver Operating Characteristic (AUROC) curves for each antibiotic classification shows that, despite the nuances, BioMegatron performs the best.}
\label{auroc}
\end{center}
\vskip -0.1in
\end{figure*}

We sourced data from the Medical Information Mart for Intensive Care IV (MIMIC-IV) and MIMIC-IV Emergency Department (ED) databases \cite{johnson2020mimic, johnson2023mimic}. This study focused on ED patients presumed to have staph infections, selected based on specific inclusion criteria. Eligible participants included those with any microbiological culture testing positive for a staph-related organism, sourced from bodily fluids such as blood, urine, cerebral spinal fluid, pleural cavity, or joint fluid, accompanied by a prescribed antibiotic whose susceptibility was subsequently tested \cite{tong2015staphylococcus, kwiecinski2020staphylococcus}. From these criteria, we identified 5976 unique prescriptions in our database. Additionally, patients with multiple ED admissions that met the criteria were analyzed separately but were grouped within the same train/test divisions to prevent test set contamination. This cohort included 10 unique antibiotics, whose prevalences are shown in Table \ref{antibiotic-prevalence-table}. A demographic overview of our cohort is presented in Appendix Section \ref{demo}.

\begin{table}[h!]
\caption{Antibiotic Prevalence in MIMIC IV Cohort}
\label{antibiotic-prevalence-table}
\vskip 0.15in
\begin{center}
\begin{small}
\begin{sc}
\begin{adjustbox}{width=\columnwidth}
\begin{tabular}{lccc}
\toprule
Antibiotic & Train & Test & Total Prevalence (\%) \\
\midrule
Clindamycin & 2645 & 624 & \textbf{54.69\%} \\
Daptomycin & 1815 & 425 & \textbf{37.51\%} \\
Erythromycin & 2626 & 639 & \textbf{54.59\%} \\
Gentamicin & 4549 & 1127 & \textbf{94.89\%} \\
Levofloxacin & 2866 & 715 & \textbf{60.00\%} \\
Oxacillin & 2702 & 667 & \textbf{56.32\%} \\
Rifampin & 1929 & 459 & \textbf{39.96\%} \\
Tetracycline & 3747 & 909 & \textbf{76.57\%} \\
Trimethoprim/sul & 3671 & 908 & \textbf{71.66\%} \\
Vancomycin & 2529 & 611 & \textbf{52.53\%} \\
\bottomrule
\end{tabular}
\end{adjustbox}
\end{sc}
\end{small}
\end{center}
\vskip -0.1in
\end{table}

To motivate our experimental setup, we examine the information available about a patient at their time of arrival in the emergency department. To predict antibiotic use, we utilize six clinical modalities from the MIMIC ED Database. These EHR modalities include arrival and triage information, medication reconciliation (medrecon), diagnostic codes (ICD-9/10), vital signs, and Pyxis data. All these data points are linked to antibiotic labels from the MIMIC database using a patient ID, visit, and Hospital Admission ID (Hadm\_id), allowing us to accurately identify the patients and their tests in which certain antibiotics were effective.

\paragraph{Experiments}
In this work, we benchmark different representation strategies of EHR to identify the most effective method for predicting antibiotic susceptibility. We approach this problem as a multilabel binary classification, where we train the same base model (Light Gradient Boosted Machines) using various representation startegies of the input. These representations include: raw tabular data; EHR-shot \cite{wornow2024ehrshot}, a foundation model for tabular EHR; and three text-based representations: word2vec, a generic language model, and a medical language model, BioMegatron \cite{shin2020biomegatron}. Additionally, we conduct a clustering of our pseudonotes using the BERTopic algorithm \cite{grootendorst2022bertopic} to determine if these embeddings can naturally cluster patients. Identifying these clusters can provide insights into their potential performance in settings like zero-shot learning and provide insights into the decision making process.

\section{Results}

\begin{figure*}[h!]
\vskip 0.1in
\begin{center}
\centerline{\includegraphics[width=6.5in]{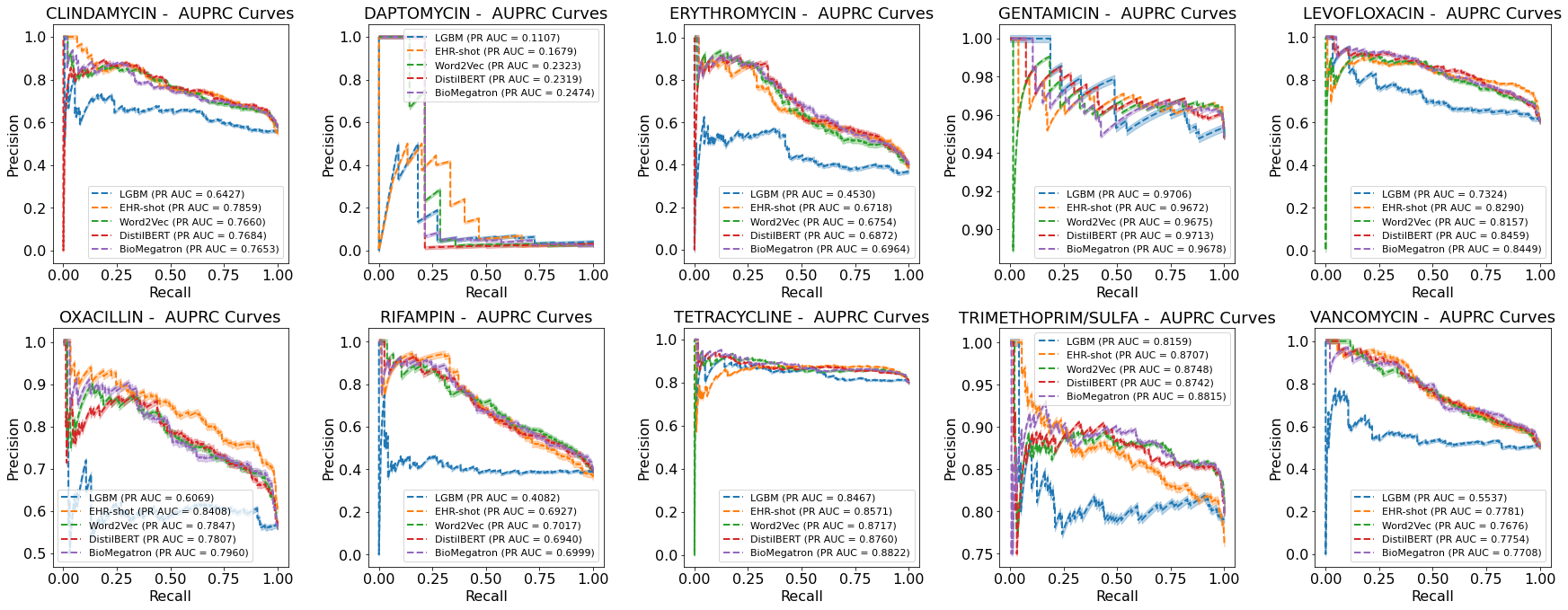}}
\caption{Area under the Precision Recall Curve (AUPRC) curves for each antibiotic classification shows that, despite the nuances, BioMegatron performs the best.}
\label{auprc}
\end{center}
\vskip -0.1in
\end{figure*}

\paragraph{Antibiotic Susceptibility Prediction} In our analysis of antibiotic prediction, we measure the Area Under the Receiver Operating Characteristic curve (AUROC) and Area Under the Precision-Recall Curve (AUPRC). Additionally, we bootstrap 1,000 times to generate 95\% confidence intervals. Our AUROC and AUPRC results are displayed in Figures \ref{auroc} and \ref{auprc}. We also measure additional F1 scores and Matthews correlation coefficients, with a whole table readout which are included in the appendix.

\paragraph{Clustering Experiment} In our clustering experiments, we aim to identify clusters using the BERTopic algorithm. By identifying clusters based on embeddings, we believe this approach can form the basis for zero-shot applications across various clinical tasks. Additionally, finding similar embeddings could provide insights into decision-making processes in these black-box models. We showcase the similarity matrix of our patient clusters in Figure \ref{sim}.

\section{Discussion}

\paragraph{Clinical Notes with Foundation Models Provide the best representation and interpretability}

\begin{figure}[t!]
\begin{center}
\centerline{\includegraphics[width=3in]{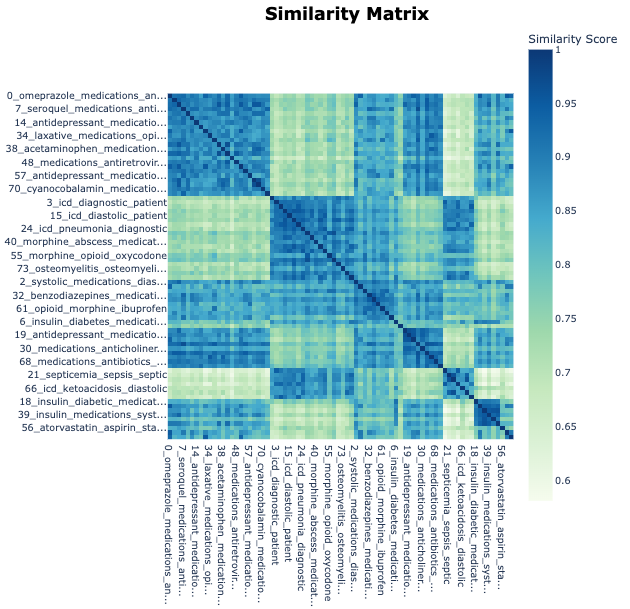}}
\caption{Identifying clusters from our patient embeddings is indicated by the squares forming along the diagonal of our similarity matrix.}
\label{sim}
\end{center}
\vspace{-1cm}
\end{figure}

From Figures \ref{auroc} and \ref{auprc}, we observe that the foundation models operating on our pseudo-notes method provide the best overall performance across most of the antibiotics. While a generic foundation model and EHR-shot excel with some antibiotics, the clinical foundation model consistently shows superior performance across both AUROC and AUPRC metrics.

Beyond enhanced predictive abilities, an advantage of our pseudo-notes method over EHR foundation models and tabular representations is its interpretability. Compared to the specific structuring required by EHR-shot, our method offers a simpler and more effective interface to understand the data that is being modeled which can improve the trust between AI and healthcare professionals.

\paragraph{Interoperability}

Another advantage of pseudo-notes over EHR foundation models is their interoperability with proprietary healthcare systems. Our method offers a straightforward interface for converting any EHR tabular data from tables to text. Our conversion also facilitates the use of off-the-shelf open-source foundation models. As improved models are continually developed, this data format provides an easy interface to adapt and swap the backbone for improved representation of our clinical text. Additionally, EHR foundation models do not operate on non-OMOP vocabularies, which limits its effectiveness on datasets like MIMIC-IV that utilize these specialized vocabularies.


\paragraph{Patient Similarity}
One final advantage of our pseudo-notes is illustrated in Figure \ref{sim}, which demonstrates the capability to perform a similarity search on our patient embeddings. From this analysis, we identified clusters related to sepsis, diabetes, stomach acid issues, anxiety, painkillers, respiratory conditions, and antidepressants. This opens up potential use cases for this data representation strategy to be used in zero-shot learning studies and offers insights into decision-making processes based on the embeddings. Further research is needed to explore both of these areas.

\section{Conclusion}

In this work, we introduced a methodology called pseudo-notes, which converts EHR tabular data into text to achieve an optimal representation strategy. We discovered that pseudo-notes outperformed various representation strategies and remains a highly flexible framework, compatible with the ongoing development of foundation model backbones, which could further enhance its performance. Additionally, we found that pseudo-notes can identify patient clusters within the EHR, opening up promising avenues for future studies in zero-shot learning and model interpretation.

From an application perspective, we demonstrated how a straightforward data transformation has emerged as an easy interface for making EHR data synergized before integrating it into machine learning models. We think that this strategy could be a better way to work with EHR data for future research and help build trust due to its interpretability. Particularly in this study, we illustrated its potential by identifying suitable antibiotics for patients arriving at the ED, where timely and accurate decisions are critical. We as a group have highlighted the importance of improving antibiotic stewardship and showcase the impact of data-driven strategies in addressing pressing healthcare challenges.

\subsection*{Impact Statement}

The goal of this work is to advance the field of Machine Learning in Healthcare, and thus presents a novel potential interface between modern NLP (Foundation models) and clinical data. 

\nocite{langley00}

\bibliography{example_paper}

\begin{thebibliography}{43}
\providecommand{\natexlab}[1]{#1}
\providecommand{\url}[1]{\texttt{#1}}
\expandafter\ifx\csname urlstyle\endcsname\relax
  \providecommand{\doi}[1]{doi: #1}\else
  \providecommand{\doi}{doi: \begingroup \urlstyle{rm}\Url}\fi

\bibitem[Agniel et~al.(2018)Agniel, Kohane, and Weber]{agniel2018biases}
Agniel, D., Kohane, I.~S., and Weber, G.~M.
\newblock Biases in electronic health record data due to processes within the healthcare system: retrospective observational study.
\newblock \emph{Bmj}, 361, 2018.

\bibitem[Alsentzer et~al.(2019)Alsentzer, Murphy, Boag, Weng, Jin, Naumann, and McDermott]{alsentzer2019publicly}
Alsentzer, E., Murphy, J.~R., Boag, W., Weng, W.-H., Jin, D., Naumann, T., and McDermott, M.
\newblock Publicly available clinical bert embeddings.
\newblock \emph{arXiv preprint arXiv:1904.03323}, 2019.

\bibitem[Bartlett et~al.(2013)Bartlett, Gilbert, and Spellberg]{bartlett2013seven}
Bartlett, J.~G., Gilbert, D.~N., and Spellberg, B.
\newblock Seven ways to preserve the miracle of antibiotics.
\newblock \emph{Clinical infectious diseases}, 56\penalty0 (10):\penalty0 1445--1450, 2013.

\bibitem[Cheng et~al.(2016)Cheng, Wang, Zhang, and Hu]{cheng2016risk}
Cheng, Y., Wang, F., Zhang, P., and Hu, J.
\newblock Risk prediction with electronic health records: A deep learning approach.
\newblock In \emph{Proceedings of the 2016 SIAM international conference on data mining}, pp.\  432--440. SIAM, 2016.

\bibitem[Churpek et~al.(2014)Churpek, Yuen, Park, Gibbons, and Edelson]{churpek2014using}
Churpek, M.~M., Yuen, T.~C., Park, S.~Y., Gibbons, R., and Edelson, D.~P.
\newblock Using electronic health record data to develop and validate a prediction model for adverse outcomes in the wards.
\newblock \emph{Critical care medicine}, 42\penalty0 (4):\penalty0 841--848, 2014.

\bibitem[Cowie et~al.(2017)Cowie, Blomster, Curtis, Duclaux, Ford, Fritz, Goldman, Janmohamed, Kreuzer, Leenay, et~al.]{cowie2017electronic}
Cowie, M.~R., Blomster, J.~I., Curtis, L.~H., Duclaux, S., Ford, I., Fritz, F., Goldman, S., Janmohamed, S., Kreuzer, J., Leenay, M., et~al.
\newblock Electronic health records to facilitate clinical research.
\newblock \emph{Clinical Research in Cardiology}, 106:\penalty0 1--9, 2017.

\bibitem[Evans(2016)]{evans2016electronic}
Evans, R.~S.
\newblock Electronic health records: then, now, and in the future.
\newblock \emph{Yearbook of medical informatics}, 25\penalty0 (S 01):\penalty0 S48--S61, 2016.

\bibitem[Ferrao et~al.(2016)Ferrao, Oliveira, Janela, and Martins]{ferrao2016preprocessing}
Ferrao, J.~C., Oliveira, M.~D., Janela, F., and Martins, H.~M.
\newblock Preprocessing structured clinical data for predictive modeling and decision support.
\newblock \emph{Applied clinical informatics}, 7\penalty0 (04):\penalty0 1135--1153, 2016.

\bibitem[Goetz et~al.(2024)Goetz, Seedat, Vandersluis, and van~der Schaar]{goetz2024generalization}
Goetz, L., Seedat, N., Vandersluis, R., and van~der Schaar, M.
\newblock Generalization—a key challenge for responsible ai in patient-facing clinical applications.
\newblock \emph{npj Digital Medicine}, 7\penalty0 (1):\penalty0 1--4, 2024.

\bibitem[Goldstein et~al.(2017)Goldstein, Navar, Pencina, and Ioannidis]{goldstein2017opportunities}
Goldstein, B.~A., Navar, A.~M., Pencina, M.~J., and Ioannidis, J.~P.
\newblock Opportunities and challenges in developing risk prediction models with electronic health records data: a systematic review.
\newblock \emph{Journal of the American Medical Informatics Association: JAMIA}, 24\penalty0 (1):\penalty0 198, 2017.

\bibitem[Golkar et~al.(2014)Golkar, Bagasra, and Pace]{golkar2014bacteriophage}
Golkar, Z., Bagasra, O., and Pace, D.~G.
\newblock Bacteriophage therapy: a potential solution for the antibiotic resistance crisis.
\newblock \emph{The Journal of Infection in Developing Countries}, 8\penalty0 (02):\penalty0 129--136, 2014.

\bibitem[Gould \& Bal(2013)Gould and Bal]{gould2013new}
Gould, I.~M. and Bal, A.~M.
\newblock New antibiotic agents in the pipeline and how they can help overcome microbial resistance.
\newblock \emph{Virulence}, 4\penalty0 (2):\penalty0 185--191, 2013.

\bibitem[Grootendorst(2022)]{grootendorst2022bertopic}
Grootendorst, M.
\newblock Bertopic: Neural topic modeling with a class-based tf-idf procedure.
\newblock \emph{arXiv preprint arXiv:2203.05794}, 2022.

\bibitem[Hegselmann et~al.(2023)Hegselmann, Buendia, Lang, Agrawal, Jiang, and Sontag]{hegselmann2023tabllm}
Hegselmann, S., Buendia, A., Lang, H., Agrawal, M., Jiang, X., and Sontag, D.
\newblock Tabllm: Few-shot classification of tabular data with large language models.
\newblock In \emph{International Conference on Artificial Intelligence and Statistics}, pp.\  5549--5581. PMLR, 2023.

\bibitem[Hoerbst \& Ammenwerth(2010)Hoerbst and Ammenwerth]{hoerbst2010electronic}
Hoerbst, A. and Ammenwerth, E.
\newblock Electronic health records.
\newblock \emph{Methods of information in medicine}, 49\penalty0 (04):\penalty0 320--336, 2010.

\bibitem[Hohman et~al.(2023)Hohman, Martinez, Klompas, Kraus, Li, Carton, Cocoros, Jackson, Karras, Wiltz, et~al.]{hohman2023leveraging}
Hohman, K.~H., Martinez, A.~K., Klompas, M., Kraus, E.~M., Li, W., Carton, T.~W., Cocoros, N.~M., Jackson, S.~L., Karras, B.~T., Wiltz, J.~L., et~al.
\newblock Leveraging electronic health record data for timely chronic disease surveillance: the multi-state ehr-based network for disease surveillance.
\newblock \emph{Journal of Public Health Management and Practice}, 29\penalty0 (2):\penalty0 162--173, 2023.

\bibitem[Johnson et~al.(2020)Johnson, Bulgarelli, Pollard, Horng, Celi, and Mark]{johnson2020mimic}
Johnson, A., Bulgarelli, L., Pollard, T., Horng, S., Celi, L.~A., and Mark, R.
\newblock Mimic-iv.
\newblock \emph{PhysioNet. Available online at: https://physionet. org/content/mimiciv/1.0/(accessed August 23, 2021)}, pp.\  49--55, 2020.

\bibitem[Johnson et~al.(2023)Johnson, Bulgarelli, Shen, Gayles, Shammout, Horng, Pollard, Hao, Moody, Gow, et~al.]{johnson2023mimic}
Johnson, A.~E., Bulgarelli, L., Shen, L., Gayles, A., Shammout, A., Horng, S., Pollard, T.~J., Hao, S., Moody, B., Gow, B., et~al.
\newblock Mimic-iv, a freely accessible electronic health record dataset.
\newblock \emph{Scientific data}, 10\penalty0 (1):\penalty0 1, 2023.

\bibitem[Kwiecinski \& Horswill(2020)Kwiecinski and Horswill]{kwiecinski2020staphylococcus}
Kwiecinski, J.~M. and Horswill, A.~R.
\newblock Staphylococcus aureus bloodstream infections: pathogenesis and regulatory mechanisms.
\newblock \emph{Current opinion in microbiology}, 53:\penalty0 51--60, 2020.

\bibitem[Lee et~al.(2020)Lee, Yoon, Kim, Kim, Kim, So, and Kang]{lee2020biobert}
Lee, J., Yoon, W., Kim, S., Kim, D., Kim, S., So, C.~H., and Kang, J.
\newblock Biobert: a pre-trained biomedical language representation model for biomedical text mining.
\newblock \emph{Bioinformatics}, 36\penalty0 (4):\penalty0 1234--1240, 2020.

\bibitem[Lee et~al.(2024)Lee, Jain, Chen, Ono, Fang, Rudas, and Chiang]{lee2024emergency}
Lee, S.~A., Jain, S., Chen, A., Ono, K., Fang, J., Rudas, A., and Chiang, J.~N.
\newblock Emergency department decision support using clinical pseudo-notes, 2024.

\bibitem[Li et~al.(2020)Li, Chen, Ritchie, and Moore]{li2020electronic}
Li, R., Chen, Y., Ritchie, M.~D., and Moore, J.~H.
\newblock Electronic health records and polygenic risk scores for predicting disease risk.
\newblock \emph{Nature Reviews Genetics}, 21\penalty0 (8):\penalty0 493--502, 2020.

\bibitem[Li et~al.(2022)Li, Mamouei, Salimi-Khorshidi, Rao, Hassaine, Canoy, Lukasiewicz, and Rahimi]{li2022hi}
Li, Y., Mamouei, M., Salimi-Khorshidi, G., Rao, S., Hassaine, A., Canoy, D., Lukasiewicz, T., and Rahimi, K.
\newblock Hi-behrt: hierarchical transformer-based model for accurate prediction of clinical events using multimodal longitudinal electronic health records.
\newblock \emph{IEEE journal of biomedical and health informatics}, 27\penalty0 (2):\penalty0 1106--1117, 2022.

\bibitem[Liu et~al.(2018)Liu, Zhang, and Razavian]{liu2018deep}
Liu, J., Zhang, Z., and Razavian, N.
\newblock Deep ehr: Chronic disease prediction using medical notes.
\newblock In \emph{Machine Learning for Healthcare Conference}, pp.\  440--464. PMLR, 2018.

\bibitem[Liu et~al.(2021)Liu, Hu, Xu, Xu, and Chen]{liu2021med}
Liu, N., Hu, Q., Xu, H., Xu, X., and Chen, M.
\newblock Med-bert: A pretraining framework for medical records named entity recognition.
\newblock \emph{IEEE Transactions on Industrial Informatics}, 18\penalty0 (8):\penalty0 5600--5608, 2021.

\bibitem[Lushniak(2014)]{lushniak2014antibiotic}
Lushniak, B.~D.
\newblock Antibiotic resistance: a public health crisis.
\newblock \emph{Public Health Reports}, 129\penalty0 (4):\penalty0 314--316, 2014.

\bibitem[Nature(2013)]{nature2013antibiotic}
Nature, E.
\newblock The antibiotic alarm.
\newblock \emph{Nature}, 495\penalty0 (7440):\penalty0 141, 2013.

\bibitem[Pang et~al.(2021)Pang, Jiang, Kalluri, Spotnitz, Chen, Perotte, and Natarajan]{pang2021cehr}
Pang, C., Jiang, X., Kalluri, K.~S., Spotnitz, M., Chen, R., Perotte, A., and Natarajan, K.
\newblock Cehr-bert: Incorporating temporal information from structured ehr data to improve prediction tasks.
\newblock In \emph{Machine Learning for Health}, pp.\  239--260. PMLR, 2021.

\bibitem[Rasmy et~al.(2021)Rasmy, Xiang, Xie, Tao, and Zhi]{rasmy2021med}
Rasmy, L., Xiang, Y., Xie, Z., Tao, C., and Zhi, D.
\newblock Med-bert: pretrained contextualized embeddings on large-scale structured electronic health records for disease prediction.
\newblock \emph{NPJ digital medicine}, 4\penalty0 (1):\penalty0 86, 2021.

\bibitem[Read \& Woods(2014)Read and Woods]{read2014antibiotic}
Read, A.~F. and Woods, R.~J.
\newblock Antibiotic resistance management.
\newblock \emph{Evolution, medicine, and public health}, 2014\penalty0 (1):\penalty0 147, 2014.

\bibitem[Sanh et~al.(2019)Sanh, Debut, Chaumond, and Wolf]{sanh2019distilbert}
Sanh, V., Debut, L., Chaumond, J., and Wolf, T.
\newblock Distilbert, a distilled version of bert: smaller, faster, cheaper and lighter.
\newblock \emph{arXiv preprint arXiv:1910.01108}, 2019.

\bibitem[Sengupta et~al.(2013)Sengupta, Chattopadhyay, and Grossart]{sengupta2013multifaceted}
Sengupta, S., Chattopadhyay, M.~K., and Grossart, H.-P.
\newblock The multifaceted roles of antibiotics and antibiotic resistance in nature.
\newblock \emph{Frontiers in microbiology}, 4:\penalty0 47, 2013.

\bibitem[Shin et~al.(2020)Shin, Zhang, Bakhturina, Puri, Patwary, Shoeybi, and Mani]{shin2020biomegatron}
Shin, H.-C., Zhang, Y., Bakhturina, E., Puri, R., Patwary, M., Shoeybi, M., and Mani, R.
\newblock Biomegatron: Larger biomedical domain language model, 2020.

\bibitem[Steinberg et~al.(2023)Steinberg, Fries, Xu, and Shah]{steinberg2023motor}
Steinberg, E., Fries, J., Xu, Y., and Shah, N.
\newblock Motor: A time-to-event foundation model for structured medical records.
\newblock \emph{arXiv preprint arXiv:2301.03150}, 2023.

\bibitem[Suter et~al.(1994)Suter, Armaganidis, Beaufils, Bonfill, Burchardi, Cook, Fagot-Largeault, Thijs, Vesconi, Williams, et~al.]{suter1994predicting}
Suter, P., Armaganidis, A., Beaufils, F., Bonfill, X., Burchardi, H., Cook, D., Fagot-Largeault, A., Thijs, L., Vesconi, S., Williams, A., et~al.
\newblock Predicting outcome in icu patients.
\newblock \emph{Intensive Care Medicine}, 20:\penalty0 390--397, 1994.

\bibitem[Sutton et~al.(2020)Sutton, Pincock, Baumgart, Sadowski, Fedorak, and Kroeker]{sutton2020overview}
Sutton, R.~T., Pincock, D., Baumgart, D.~C., Sadowski, D.~C., Fedorak, R.~N., and Kroeker, K.~I.
\newblock An overview of clinical decision support systems: benefits, risks, and strategies for success.
\newblock \emph{NPJ digital medicine}, 3\penalty0 (1):\penalty0 17, 2020.

\bibitem[Tang et~al.(2020)Tang, Davarmanesh, Song, Koutra, Sjoding, and Wiens]{tang2020democratizing}
Tang, S., Davarmanesh, P., Song, Y., Koutra, D., Sjoding, M.~W., and Wiens, J.
\newblock Democratizing ehr analyses with fiddle: a flexible data-driven preprocessing pipeline for structured clinical data.
\newblock \emph{Journal of the American Medical Informatics Association}, 27\penalty0 (12):\penalty0 1921--1934, 2020.

\bibitem[Tong et~al.(2015)Tong, Davis, Eichenberger, Holland, and Fowler~Jr]{tong2015staphylococcus}
Tong, S.~Y., Davis, J.~S., Eichenberger, E., Holland, T.~L., and Fowler~Jr, V.~G.
\newblock Staphylococcus aureus infections: epidemiology, pathophysiology, clinical manifestations, and management.
\newblock \emph{Clinical microbiology reviews}, 28\penalty0 (3):\penalty0 603--661, 2015.

\bibitem[Ventola(2015)]{ventola2015antibiotic}
Ventola, C.~L.
\newblock The antibiotic resistance crisis: part 1: causes and threats.
\newblock \emph{Pharmacy and therapeutics}, 40\penalty0 (4):\penalty0 277, 2015.

\bibitem[Viswanathan(2014)]{viswanathan2014off}
Viswanathan, V.
\newblock Off-label abuse of antibiotics by bacteria.
\newblock \emph{Gut microbes}, 5\penalty0 (1):\penalty0 3--4, 2014.

\bibitem[Wolf et~al.(2019)Wolf, Debut, Sanh, Chaumond, Delangue, Moi, Cistac, Rault, Louf, Funtowicz, et~al.]{wolf2019huggingface}
Wolf, T., Debut, L., Sanh, V., Chaumond, J., Delangue, C., Moi, A., Cistac, P., Rault, T., Louf, R., Funtowicz, M., et~al.
\newblock Huggingface's transformers: State-of-the-art natural language processing.
\newblock \emph{arXiv preprint arXiv:1910.03771}, 2019.

\bibitem[Wornow et~al.(2024)Wornow, Thapa, Steinberg, Fries, and Shah]{wornow2024ehrshot}
Wornow, M., Thapa, R., Steinberg, E., Fries, J., and Shah, N.
\newblock Ehrshot: An ehr benchmark for few-shot evaluation of foundation models.
\newblock \emph{Advances in Neural Information Processing Systems}, 36, 2024.

\bibitem[Wu et~al.(2010)Wu, Roy, and Stewart]{wu2010prediction}
Wu, J., Roy, J., and Stewart, W.~F.
\newblock Prediction modeling using ehr data: challenges, strategies, and a comparison of machine learning approaches.
\newblock \emph{Medical care}, 48\penalty0 (6):\penalty0 S106--S113, 2010.

\end{thebibliography}
\bibliographystyle{icml2024}

\newpage
\appendix
\onecolumn
\section{Appendix}

\subsection{Additional Commentary}

\paragraph{Limitations} Some limitations of this work include the variability in patients' histories and the 512 sequence length limitation imposed by the DistilBERT \cite{sanh2019distilbert} and BioMegatron models \cite{shin2020biomegatron}. Consequently, portions of a patient's medical history may be truncated depending on the length of that history. Tokenization strategies (e.g., sub-word tokenization) can significantly influence how we handle the analysis.

\paragraph{Future Work} Future work in our group, from a methodological perspective, aims to explore how these notes can enhance studies in model interpretability and zero-shot or few-shot frameworks. From an application standpoint, we are interested in applying this methodology across various departments and applications. We plan to collaborate with clinicians throughout our institution to determine the types of clinical decision support models that are most needed and to assess how AI can benefit these healthcare facilities. 

Additionally, future work will include benchmarking the plethora of foundation models available on the Huggingface Platform \cite{wolf2019huggingface}. This will help us identify the best foundation model for specific tasks and determine whether these embeddings are task-agnostic.

\section{Dataset Characteristics}
\label{demo}

\subsection{Patient Demographics}

\begin{table}[h!]
\caption{MIMIC IV Cohort Data Overview}
\label{mimic-iv-table}
\vskip 0.15in
\begin{center}
\begin{small}
\begin{sc}
\begin{tabular}{lcccc}
\toprule
Description & Category & Train & Test & Totals \\
\midrule
Prescription, n & Total & 4803 & 1173 & 5976 \\
Unique ID, n & Total & 3283 & 878 & 4161 \\
Age Mean (SD) & & 59 (17) & 58 (17) & \\
Sex \% & Female & 1341 & 351 & 1692 \\
& Male & 1942 & 527 & 2469 \\
Race/Ethnicity \% & White & 2212 & 583 & 2795 \\
& Black & 416 & 119 & 535 \\
& Other & 401 & 96 & 497 \\
& Hispanic/Latino & 150 & 55 & 205 \\
& Asian & 88 & 20 & 108 \\
& Unable & 12 & 3 & 15 \\
& Native Hawaiian & 4 & 2 & 6 \\
\bottomrule
\end{tabular}
\end{sc}
\end{small}
\end{center}
\vskip -0.1in
\end{table}

\subsection{Clinical Modalities}

\begin{table}[h!]
\caption{Overview of Clinical Modalities in Emergency Department Visits}
\label{clinical-modalities-table}
\begin{center}
\begin{small}
\begin{tabularx}{5in}{lX} 
\toprule
Modality Name & Description \\
\midrule
Arrival Information & Records patient demographics, time of arrival, and mode of arrival (e.g., ambulance, walk-in). \\
Triage Information & Documents vital signs, severity of condition using scales like ESI, and initial chief complaints upon arrival. \\
Medication Reconciliation & Details previous and current medications the patient is taking, including dosages and frequency. \\
Patient Vitals & Ongoing measurements throughout the ED visit including heart rate, blood pressure, temperature, etc. \\
Diagnosis Codes & ICD-9/10 codes used to classify and record diagnoses during the visit. \\
Pyxis Information & Information on medications administered during the ED stay via the Pyxis system, including timing and dosage. \\
\bottomrule
\end{tabularx}
\end{small}
\end{center}
\end{table}

\newpage

\section{Results}

\begin{table}[h!]
\caption{Performance Metrics for Clindamycin}
\label{table-clindamycin}
\vskip 0.15in
\begin{center}
\begin{small}
\begin{sc}
\begin{adjustbox}{width=5in}
\begin{tabular}{l|ccccc}
\toprule
Metric & Tabular & EHR-shot & Word2Vec & DistilBERT & BioMegatron \\
\midrule
F1 & 0.7179 $\pm$ 0.032 & 0.7719 $\pm$ 0.019 & 0.7737 $\pm$ 0.031 & \textbf{0.7786 $\pm$ 0.015} & 0.7689 $\pm$ 0.029 \\
MCC & 0.0914 $\pm$ 0.011 & \textbf{0.4162 $\pm$ 0.0624} & 0.3561 $\pm$ 0.026 & 0.3772 $\pm$ 0.028 & 0.3379 $\pm$ 0.022 \\
ROC-AUC & 0.6029 $\pm$ 0.044 & \textbf{0.7664 $\pm$ 0.020} & 0.7263 $\pm$ 0.023 & 0.7443 $\pm$ 0.030 & 0.7244 $\pm$ 0.034 \\
PRC-AUC & 0.6427 $\pm$ 0.010 & \textbf{0.7859 $\pm$ 0.026} & 0.7660 $\pm$ 0.013 & 0.7684 $\pm$ 0.015 & 0.7653 $\pm$ 0.019 \\
\bottomrule
\end{tabular}
\end{adjustbox}
\end{sc}
\end{small}
\end{center}
\vskip -0.1in
\end{table}


\begin{table}[h!]
\caption{Performance Metrics for Daptomycin}
\label{table-daptomycin}
\vskip 0.15in
\begin{center}
\begin{small}
\begin{sc}
\begin{adjustbox}{width=5in}
\begin{tabular}{l| ccccc}
\toprule
Metric & Tabular & EHR-shot & Word2Vec & DistilBERT & BioMegatron \\
\midrule
F1 & 0.2667 $\pm$ 0.032 & \textbf{0.3704 $\pm$ 0.069} & 0.3333 $\pm$ 0.050 & 0.3529 $\pm$ 0.012 & 0.3529 $\pm$ 0.035 \\
MCC & 0.2867 $\pm$ 0.065 & 0.3584 $\pm$ 0.022 & 0.3943 $\pm$ 0.041 & 0.4586 $\pm$ 0.058 & \textbf{0.4587 $\pm$ 0.034} \\
ROC-AUC & 0.6107 $\pm$ 0.063 & 0.6651 $\pm$ 0.065 & 0.60223 $\pm$ 0.070 & 0.5211 $\pm$ 0.060 & \textbf{0.6708 $\pm$ 0.062} \\
PRC-AUC & 0.1107 $\pm$ 0.006 & 0.1679 $\pm$ 0.015 & 0.2323 $\pm$ 0.004 & 0.2319 $\pm$ 0.005 & \textbf{0.2474 $\pm$ 0.006} \\
\bottomrule
\end{tabular}
\end{adjustbox}
\end{sc}
\end{small}
\end{center}
\vskip -0.1in
\end{table}


\begin{table}[h!]
\caption{Performance Metrics for Erythromycin}
\label{table-erythromycin}
\vskip 0.15in
\begin{center}
\begin{small}
\begin{sc}
\begin{adjustbox}{width=5in}
\begin{tabular}{l| ccccc}
\toprule
Metric & Tabular & EHR-shot & Word2Vec & DistilBERT & BioMegatron \\
\midrule
F1 & 0.5495 $\pm$ 0.030 & 0.6575 $\pm$ 0.023 & 0.6394 $\pm$ 0.038 & \textbf{0.6592 $\pm$ 0.020} & 0.6473 $\pm$ 0.025 \\
MCC & 0.1306 $\pm$ 0.021 & \textbf{0.3807 $\pm$ 0.029} & 0.3209 $\pm$ 0.042 & 0.3702 $\pm$ 0.037 & 0.3406 $\pm$ 0.028 \\
ROC-AUC & 0.5879 $\pm$ 0.044 & 0.7590 $\pm$ 0.022 & 0.7320 $\pm$ 0.025 & 0.7597 $\pm$ 0.023 & \textbf{0.7600 $\pm$ 0.025} \\
PRC-AUC & 0.4530 $\pm$ 0.017 & 0.6718 $\pm$ 0.024 & 0.6754 $\pm$ 0.016 & 0.6872 $\pm$ 0.012 & \textbf{0.6964 $\pm$ 0.014} \\
\bottomrule
\end{tabular}
\end{adjustbox}
\end{sc}
\end{small}
\end{center}
\vskip -0.1in
\end{table}

\begin{table}[h!]
\caption{Performance Metrics for Gentamicin}
\label{table-gentamicin}
\vskip 0.15in
\begin{center}
\begin{small}
\begin{sc}
\begin{adjustbox}{width=5in}
\begin{tabular}{l| ccccc}
\toprule
Metric & Tabular & EHR-shot & Word2Vec & DistilBERT & BioMegatron \\
\midrule
F1 & 0.9762 $\pm$ 0.030 & 0.9775 $\pm$ 0.065 & \textbf{0.9776 $\pm$ 0.040} & 0.9766 $\pm$ 0.045 & \textbf{0.9776 $\pm$ 0.032} \\
MCC & 0.2521 $\pm$ 0.055 & 0.3634 $\pm$ 0.021 & 0.3953 $\pm$ 0.030 & \textbf{0.3969 $\pm$ 0.065} & 0.3667 $\pm$ 0.035 \\
ROC-AUC & 0.6158 $\pm$ 0.089 & 0.6310 $\pm$ 0.047 & 0.6727 $\pm$ 0.047 & \textbf{0.6777 $\pm$ 0.042} & 0.6523 $\pm$ 0.039 \\
PRC-AUC & 0.9706 $\pm$ 0.036 & 0.9672 $\pm$ 0.004 & 0.9675 $\pm$ 0.002 & \textbf{0.9713 $\pm$ 0.002} & 0.9678 $\pm$ 0.011 \\
\bottomrule
\end{tabular}
\end{adjustbox}
\end{sc}
\end{small}
\end{center}
\vskip -0.1in
\end{table}


\begin{table}[h!]
\caption{Performance Metrics for Levofloxacin}
\label{table-levofloxacin}
\vskip 0.15in
\begin{center}
\begin{small}
\begin{sc}
\begin{adjustbox}{width=5in}
\begin{tabular}{l| ccccc}
\toprule
Metric & Tabular & EHR-shot & Word2Vec & DistilBERT & BioMegatron \\
\midrule
F1 & 0.7641 $\pm$ 0.028 & \textbf{0.8386 $\pm$ 0.017} & 0.8088 $\pm$ 0.012 & 0.8034 $\pm$ 0.013 & 0.8066 $\pm$ 0.013 \\
MCC & 0.1766 $\pm$ 0.025 & \textbf{0.5094 $\pm$ 0.015} & 0.4302 $\pm$ 0.025 & 0.4260 $\pm$ 0.017 & 0.4261 $\pm$ 0.017 \\
ROC-AUC & 0.6326 $\pm$ 0.034 & 0.7972 $\pm$ 0.017 & 0.7787 $\pm$ 0.021 & \textbf{0.7974 $\pm$ 0.018} & 0.7937 $\pm$ 0.021 \\
PRC-AUC & 0.7324 $\pm$ 0.013 & 0.8290 $\pm$ 0.014 & 0.8157 $\pm$ 0.011 & \textbf{0.8459 $\pm$ 0.012} & 0.8449 $\pm$ 0.014 \\
\bottomrule
\end{tabular}
\end{adjustbox}
\end{sc}
\end{small}
\end{center}
\vskip -0.1in
\end{table}

\begin{table}[h!]
\caption{Performance Metrics for Oxacillin}
\label{table-oxacillin}
\vskip 0.15in
\begin{center}
\begin{small}
\begin{sc}
\begin{adjustbox}{width=5in}
\begin{tabular}{l| ccccc}
\toprule
Metric & Tabular & EHR-shot & Word2Vec & DistilBERT & BioMegatron \\
\midrule
F1 & 0.7264 $\pm$ 0.027 & \textbf{0.8229 $\pm$ 0.024} & 0.7899 $\pm$ 0.018 & 0.7790 $\pm$ 0.023 & 0.7975 $\pm$ 0.014 \\
MCC & 0.2012 $\pm$ 0.015 & \textbf{0.4955 $\pm$ 0.017} & 0.4456 $\pm$ 0.021 & 0.4028 $\pm$ 0.020 & 0.4674 $\pm$ 0.018 \\
ROC-AUC & 0.5607 $\pm$ 0.027 & \textbf{0.7996 $\pm$ 0.016} & 0.7688 $\pm$ 0.017 & 0.7692 $\pm$ 0.013 & 0.7723 $\pm$ 0.015 \\
PRC-AUC & 0.6069 $\pm$ 0.011 & \textbf{0.8408 $\pm$ 0.018} & 0.7847 $\pm$ 0.019 & 0.7807 $\pm$ 0.018 & 0.7960 $\pm$ 0.017 \\
\bottomrule
\end{tabular}
\end{adjustbox}
\end{sc}
\end{small}
\end{center}
\vskip -0.1in
\end{table}

\begin{table}[h!]
\caption{Performance Metrics for Rifampin}
\label{table-rifampin}
\vskip 0.15in
\begin{center}
\begin{small}
\begin{sc}
\begin{adjustbox}{width=5in}
\begin{tabular}{l| ccccc}
\toprule
Metric & Tabular & EHR-shot & Word2Vec & DistilBERT & BioMegatron \\
\midrule
F1 & 0.5619 $\pm$ 0.026 & 0.6250 $\pm$ 0.024 & \textbf{0.6582 $\pm$ 0.012} & 0.6455 $\pm$ 0.017 & 0.6434 $\pm$ 0.018 \\
MCC & $\geq$0.0000 $\pm$ 0.000 & \textbf{0.4136 $\pm$ 0.027} & 0.3907 $\pm$ 0.012 & 0.3599 $\pm$ 0.017 & 0.3583 $\pm$ 0.021 \\
ROC-AUC & 0.5083 $\pm$ 0.026 & 0.7634 $\pm$ 0.015 & \textbf{0.7691 $\pm$ 0.015} & 0.7553 $\pm$ 0.016 & 0.7644 $\pm$ 0.015 \\
PRC-AUC & 0.4082 $\pm$ 0.002 & 0.6927 $\pm$ 0.011 & \textbf{0.7017 $\pm$ 0.013} & 0.6940 $\pm$ 0.011 & 0.6999 $\pm$ 0.012 \\
\bottomrule
\end{tabular}
\end{adjustbox}
\end{sc}
\end{small}
\end{center}
\vskip -0.1in
\end{table}

\begin{table}[h!]
\caption{Performance Metrics for Tetracycline}
\label{table-tetracycline}
\vskip 0.15in
\begin{center}
\begin{small}
\begin{sc}
\begin{adjustbox}{width=5in}
\begin{tabular}{l| ccccc}
\toprule
Metric & Tabular & EHR-shot & Word2Vec & DistilBERT & BioMegatron \\
\midrule
F1 & 0.8950 $\pm$ 0.025 & 0.9009 $\pm$ 0.024 & 0.9028 $\pm$ 0.027 & 0.9035 $\pm$ 0.023 & \textbf{0.9049 $\pm$ 0.021} \\
MCC & 0.1657 $\pm$ 0.025 & 0.3805 $\pm$ 0.012 & 0.3696 $\pm$ 0.015 & 0.3795 $\pm$ 0.017 & \textbf{0.3865 $\pm$ 0.021} \\
ROC-AUC & 0.5822 $\pm$ 0.035 & 0.6908 $\pm$ 0.018 & 0.6843 $\pm$ 0.023 & 0.6843 $\pm$ 0.025 & \textbf{0.6933 $\pm$ 0.023} \\
PRC-AUC & 0.8467 $\pm$ 0.004 & 0.8571 $\pm$ 0.005 & 0.8717 $\pm$ 0.005 & 0.8760 $\pm$ 0.003 & \textbf{0.8822 $\pm$ 0.002} \\
\bottomrule
\end{tabular}
\end{adjustbox}
\end{sc}
\end{small}
\end{center}
\vskip -0.1in
\end{table}

\begin{table}[h!]
\caption{Performance Metrics for Trimethoprim/sulfa}
\label{table-trimethoprim-sulfa}
\vskip 0.15in
\begin{center}
\begin{small}
\begin{sc}
\begin{adjustbox}{width=5in}
\begin{tabular}{l| ccccc}
\toprule
Metric & Tabular & EHR-shot & Word2Vec & DistilBERT & BioMegatron \\
\midrule
F1 & 0.8835 $\pm$ 0.018 & 0.8856 $\pm$ 0.027 & 0.9080 $\pm$ 0.032 & 0.9080 $\pm$ 0.024 & \textbf{0.9100 $\pm$ 0.025} \\
MCC & $\geq$ 0.0000 $\pm$ 0.000 & 0.3785 $\pm$ 0.023 & 0.4070 $\pm$ 0.031 & 0.4162 $\pm$ 0.023 & \textbf{0.4321 $\pm$ 0.032} \\
ROC-AUC & 0.5393 $\pm$ 0.031 & 0.7026 $\pm$ 0.016 & 0.7025 $\pm$ 0.018 & 0.6946 $\pm$ 0.027 & \textbf{0.7122 $\pm$ 0.026} \\
PRC-AUC & 0.8159 $\pm$ 0.017 & 0.8707 $\pm$ 0.015 & 0.8748 $\pm$ 0.004 & 0.8742 $\pm$ 0.004 & \textbf{0.8815 $\pm$ 0.008} \\
\bottomrule
\end{tabular}
\end{adjustbox}
\end{sc}
\end{small}
\end{center}
\vskip -0.1in
\end{table}

\begin{table}[h!]
\caption{Performance Metrics for Vancomycin}
\label{table-vancomycin}
\vskip 0.15in
\begin{center}
\begin{small}
\begin{sc}
\begin{adjustbox}{width=5in}
\begin{tabular}{l| ccccc}
\toprule
Metric & Tabular & EHR-shot & Word2Vec & DistilBERT & BioMegatron \\
\midrule
F1 & 0.6786 $\pm$ 0.021 & 0.7201 $\pm$ 0.016 & 0.7227 $\pm$ 0.015 & 0.7244 $\pm$ 0.014 & \textbf{0.7287 $\pm$ 0.023} \\
MCC & 0.1433 $\pm$ 0.026 & 0.3342 $\pm$ 0.023 & 0.3382 $\pm$ 0.026 & 0.3370 $\pm$ 0.025 & \textbf{0.3555 $\pm$ 0.024} \\
ROC-AUC & 0.5431 $\pm$ 0.020 & 0.7566 $\pm$ 0.014 & 0.7449 $\pm$ 0.011 & 0.7542 $\pm$ 0.012 & \textbf{0.7610 $\pm$ 0.013} \\
PRC-AUC & 0.5537 $\pm$ 0.005 & \textbf{0.7781 $\pm$ 0.018} & 0.7676 $\pm$ 0.015 & 0.7754 $\pm$ 0.019 & 0.7708 $\pm$ 0.006 \\
\bottomrule
\end{tabular}
\end{adjustbox}
\end{sc}
\end{small}
\end{center}
\vskip -0.1in
\end{table}

\begin{table}[h!]
\caption{Number of Winning Metrics}
\label{table-daptomycin}
\vskip 0.15in
\begin{center}
\begin{small}
\begin{sc}
\begin{adjustbox}{width=5in}
\begin{tabular}{l| ccccc| c}
\toprule
Metric & Tabular & EHR-shot & Word2Vec & DistilBERT & BioMegatron & Total \\
\midrule
Number & 0 & 13 & 3 & 7 &\textbf{17} & 40\\
\bottomrule
\end{tabular}
\end{adjustbox}
\end{sc}
\end{small}
\end{center}
\vskip -0.1in
\end{table}

\end{document}